\documentclass[10pt, a4paper]{article}

\usepackage[final]{lrec2026} % this is the new style
\usepackage{comment}
\usepackage{url}
\usepackage{float}
\usepackage{csquotes}
\usepackage{amssymb} 
\usepackage{moresize}
\usepackage{multirow,graphicx,booktabs,makecell,array}

\title{Analysing Calls to Order in German Parliamentary Debates}

\name{Nina Smirnova\textsuperscript{1}, Daniel Dan\textsuperscript{2}, Philipp Mayr\textsuperscript{1}} 

\address{\textsuperscript{1} GESIS – Leibniz Institute for the Social Sciences, Germany\\
\textsuperscript{2} Modul University Vienna, Austria \\
         \{nina.smirnova, philipp.mayr\}@gesis.org, daniel.dan@modul.ac.at}

\abstract{
Parliamentary debate constitutes a central arena of political power, shaping legislative outcomes and public discourse. Incivility within this arena signals political polarization and institutional conflict. This study presents a systematic investigation of incivility in the German Bundestag by examining calls to order (CtO; plural: CtOs) as formal indicators of norm violations. Despite their relevance, CtOs have received little systematic attention in parliamentary research.
We introduce a rule-based method for detecting and annotating CtOs in parliamentary speeches and present a novel dataset of German parliamentary debates spanning 72 years that includes annotated CtO instances. Additionally, we develop the first classification system for CtO triggers and analyze the factors associated with their occurrence. Our findings show that, despite formal regulations, the issuance of CtOs is partly subjective and influenced by session presidents and parliamentary dynamics, with certain individuals disproportionately affected. An insult towards individuals is the most frequent cause of CtO. In general, male members and those belonging to opposition parties receive more calls to order than their female and coalition-party counterparts. Most CtO triggers were detected in speeches dedicated to governmental affairs and actions of the presidency. The CtO triggers dataset is available at: 
\url{https://github.com/kalawinka/cto_analysis}.
 \\ \newline  }

\begin{document}

\maketitleabstract

\section{Introduction}

%Parliamentary debate analysis can reveal political parties' hidden programmatic-ideological positions. From the linguistic perspective, parliamentary debates were analysed in terms of political discourse \citep{Dijk2004OnTA}, sentiment and position-taking \citep{abercrombie_sentiment_2020}, and cross-cultural \citep{bayley_cross-cultural_2004} and gender differences \citep{stecker_evolution_2021, ash_gender_2024}. 

Parliamentary debate constitutes a central arena of political power, shaping legislative outcomes and influencing the tone of public discourse. The presence of incivility within parliamentary proceedings serves as an important indicator of political polarization and institutional conflict. This study presents a systematic investigation of incivility in parliamentary discourse, focusing on calls to order (CtOs) in the German Bundestag.
The call to order (CtO) serves as a valuable instrument for examining negativity and incivility in political debates and provides a unique perspective on political polarization \citep{Jenny_Marcelo_2021}. Moreover, analyzing calls to order (CtOs) as markers of disruptive language represents a novel approach to the study of parliamentary corpora, extending beyond traditional sentiment or stance analysis. Furthermore, methods of automatic analysis applied to parliamentary data support government transparency and accountability.
However, a notable gap exists in the analysis of CtOs in parliamentary discourse. To the best of our knowledge, the sole effort in this area is that of  \citep{Jenny_Marcelo_2021}. 
In this study, we present a novel and comprehensive analysis of speeches delivered by German politicians over 72 years of parliamentary history, employing both automated and manual methods. CtOs have been largely overlooked in political research. Consequently, this paper represents the first attempt to develop a classification system for the triggers of CtOs (Table~\ref{tab:classification}) and to analyze the factors contributing to incivility in parliamentary discourse. Moreover, we propose a rule-based method for the detection and annotation of CtOs within parliamentary speeches and introduce a novel dataset comprising annotated speeches that include a CtO.

In the present research, we will address the following research questions: \textbf{RQ1:} Which topics triggered the highest number of CtOs?
\textbf{RQ2:} What are the most frequent classes of triggers for issuing a CtO?
\textbf{RQ3:} How do factors such as political party affiliation, individual politicians, legislative periods, and debate topics relate to the issuance of CtOs?

\subsection*{Terminology used in this paper}\label{sec:terms}

\begin{figure}[h]
  \centering
  \includegraphics[width=1\linewidth]{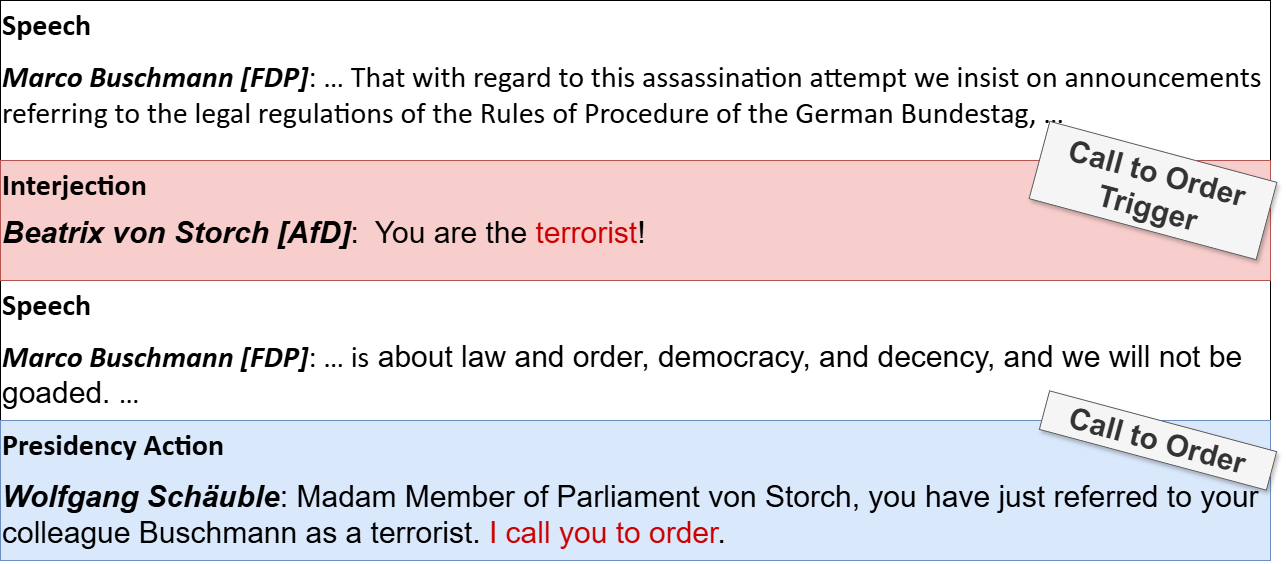}
  \caption{Example of a trigger (red) and an issued call to order (blue). Translated to English from German debates.}
  \label{fig:cto_example}
\end{figure}

A call to order, issued by the president of the session, serves as a disciplinary measure in response to breaches of parliamentary protocol, such as instances of personal insults among members or disruptions to the proceedings. Only the president may call members of the German parliament to order by stating their name \cite[p.~447]{schindler_datenhandbuch_2005}. 
Figure~\ref{fig:cto_example} demonstrates an example of CtO and how it is triggered during a parliamentary session. In the present paper, speeches of the president are referred to as presidency actions. An interjection is an interruption during a speech or introduction of another person\footnote{\url{https://de.wiktionary.org/wiki/Zwischenruf}}.
%\footnote{\url{https://www.bundestag.de/resource/blob/196296/4b2ee134475f75e677cdf679caff93a8/Kapitel_07_16_Ordnungsma__nahmen.pdf}}

A legislative period (LP) is a period in which a parliament can act as a lawmaker and generally lasts four years in Germany. Our data spans a period from September 7, 1949, to September 7, 2021, which covers 19 legislative periods (LPs).

\section{Related work}\label{sec:litr}

%gender
Recent research on parliamentary discourse has focused on the use of automated or semi-automated analytical methods. Within the framework of gender-based research, \citet{ash_gender_2024} examined the differences between reactions to speeches given by male and female parliamentary members (PM) in the German parliament, focusing on interruptions and employing topic modelling techniques. Similarly, \citet{mandravickaite-krilavicius-2017-stylometric} investigated gender differences in language use in the professional environment based on parliamentary speeches in the Lithuanian Parliament using stylometric analysis. In the context of the United States, \citet{MILLER_SUTHERLAND_2023} analyzed interruptions in congressional hearings to explore interruption behavior influenced by gender and topic. 

%sentiment
In the domain of sentiment analysis, \citet{abercrombie-batista-navarro-2020-parlvote} introduced ParlVote, a benchmark corpus designed for the evaluation of sentiment analysis methods in the political domain, utilizing transcripts from the UK House of Commons debates. Several experimental approaches were applied to assess sentiment analysis performance on this dataset. Additionally, \citet{tarkka-etal-2024-automated} compared the performance of generative (GPT) and fine-tuned BERT-based models in emotion detection tasks applied to transcripts of Finnish parliamentary plenary sessions.

%framing
Within the scope of discursive framing research, \citet{reinig-etal-2024-politics} analyzed speech acts in German parliamentary debates using a manually annotated dataset in combination with a fine-tuned BERT-based classifier. In a related effort, \citet{rehbein-etal-2024-mouths} examined the use of factive expressions in political rhetoric and introduced GePaDe\_SpkAtt, a corpus for speaker attribution based on the German parliamentary debates. This work also involved training a model for predicting speech events across a large corpus of parliamentary texts.  

%negativity
From a perspective of negativity analysis, \citet{Jenny_Marcelo_2021} analysed negativity in Austrian parliamentary discourse by predicting instances of calls to order. Further, \citet{haselmayer_how_2022} explored whether the speaker's gender and debate context impact the level of negativity, utilizing sentiment analysis and word embedding techniques.

%German parliamentary debates are widely used in political science research \cite{blaette-etal-2020-europeanization,hilmar_poison_2022,stecker_evolution_2021, ash_gender_2024}. Transcripts of speeches are openly available through the Open Data service of German parliament\footnote{\url{https://www.bundestag.de/services/opendata}}. Additionally, various corpora are available, such as GermaParl \cite{Blaette2017}, Open Discourse project\footnote{\url{https://opendiscourse.de/}}, polit-X\footnote{\url{https://polit-x.de/en/}} and others.

\section{Data and Method}\label{sec:ata_method}
We utilized an annotated XML version of the GermaParl corpus \citeplanguageresource{Blaette2017}, which comprises a collection of transcribed protocols of debates in the German parliament. The raw data underwent processing, including conversion to a format optimized for analysis, splitting speech contributions into sentences and explicit parsing sentences containing CtOs. Data processing workflow and rules-based procedure for matching sentences containing a CtO are described in detail in \citet{smirnova_open_2025}. All speeches used for the analysis can be retrieved from the PoliCorp portal\footnote{\url{https://policorp.pollux-fid.de/}}. Calls to order in the German parliament are regulated; consequently, specific words indicating a CtO are used. Therefore, we employed a rule-based approach to identify CtOs within parliamentary speeches. Two text-matching rules were applied for identifying calls to order.
Rule 1 detects occurrences of “ordnungsruf” (call to order) when it appears with “erteile(n)” (to issue / to give) but is not preceded by negations such as “kein(en)” (no / none) or “erteilten” (issued, past tense), nor accompanied by “nicht” (not).
Rule 2 matches the phrase “zur Ordnung” (to order) only when “rufe” (call) is present and it is not preceded by “Gesetz” (law) or “Gesetzes” (of the law). Rules were developed based on a manual review of a subset of the dataset containing only the speeches of the session's president. As Table~\ref{tab:corpus_counts} demonstrates, 42\% of all speech contributions in GermaParl are presidency actions, and 0,1\% of presidency actions contain a CtO.

\begin{table}[H]
\scriptsize
\centering
    \begin{tabular}{ll}
    \hline
       & count \\
    \hline
    total speech contributions  & 958,098\\
    presidency actions &  399,807\\
    speech contributions containing a call to order & 558\\
    \hline
    \end{tabular}
    \caption{Number of speech contributions and calls to order in GermaParl corpus}
    \label{tab:corpus_counts}
\end{table}

Subsequently, we extracted references to individuals mentioned in these calls using a Named Entity Recognition (NER) model \cite{akbik2018coling}, trained to recognize 4 types of entities in German texts, including names of individuals. CtOs that lacked identifiable individuals or referenced multiple individuals were manually annotated. Finally, we applied a rule-based method to resolve ambiguities among identified individuals and match them with a comprehensive database of all members of the German parliament since 1949\footnote{\url{https://www.bundestag.de/services/opendata}}. We could disambiguate persons called to order in 88\% of the speeches containing a CtO (Appendix~\ref{app:data_stat}, Table~\ref{tab:corpus_stat_app}).

As the last step, we analysed and manually annotated speech contributions containing a CtO, categorizing them according to the underlying cause that triggered the CtO. As no prior classification existed, a schema based on manual analysis of CtOs was developed and is shown in Table~\ref{tab:classification}\footnote{A detailed classification schema can be found in Appendix~\ref{app:annot_schema} (Table~\ref{tab:classification_ext}).}. Additionally, a classification model trained to detect 21 parliamentary topics \cite{klamm2022frameast} was applied to each speech, excluding those of the sessions' president, as they primarily consisted of procedural moderation and were coded as presidency action.

\begin{table}[h]
\scriptsize
    \centering
    \begin{tabular}{p{.12\textwidth}p{.3\textwidth}}
    \hline
        \textbf{class name} & \textbf{description} \\
    \hline    
    insult towards individual (\textbf{ITO}) & insult towards an individual \\
    \hline
    general insult (\textbf{GI}) & insult towards a group of people, party, event, actions, etc.\\
    \hline
    non-verbal (\textbf{NV})  & non-verbal actions that caused a call to order \\
    \hline
    not documented verbal (\textbf{NDV}) & verbal actions that caused a call to order but were not transcribed \\
    \hline
    miscellaneous (\textbf{MISC}) &  all other verbal actions excluding direct insults that caused a call to order \\
    \hline
    \end{tabular}
    \caption{Classification schema.}
    \label{tab:classification}
\end{table}

\section{Results}\label{sec:results}

\begin{figure}[h]
  \centering
  \includegraphics[width=1\linewidth]{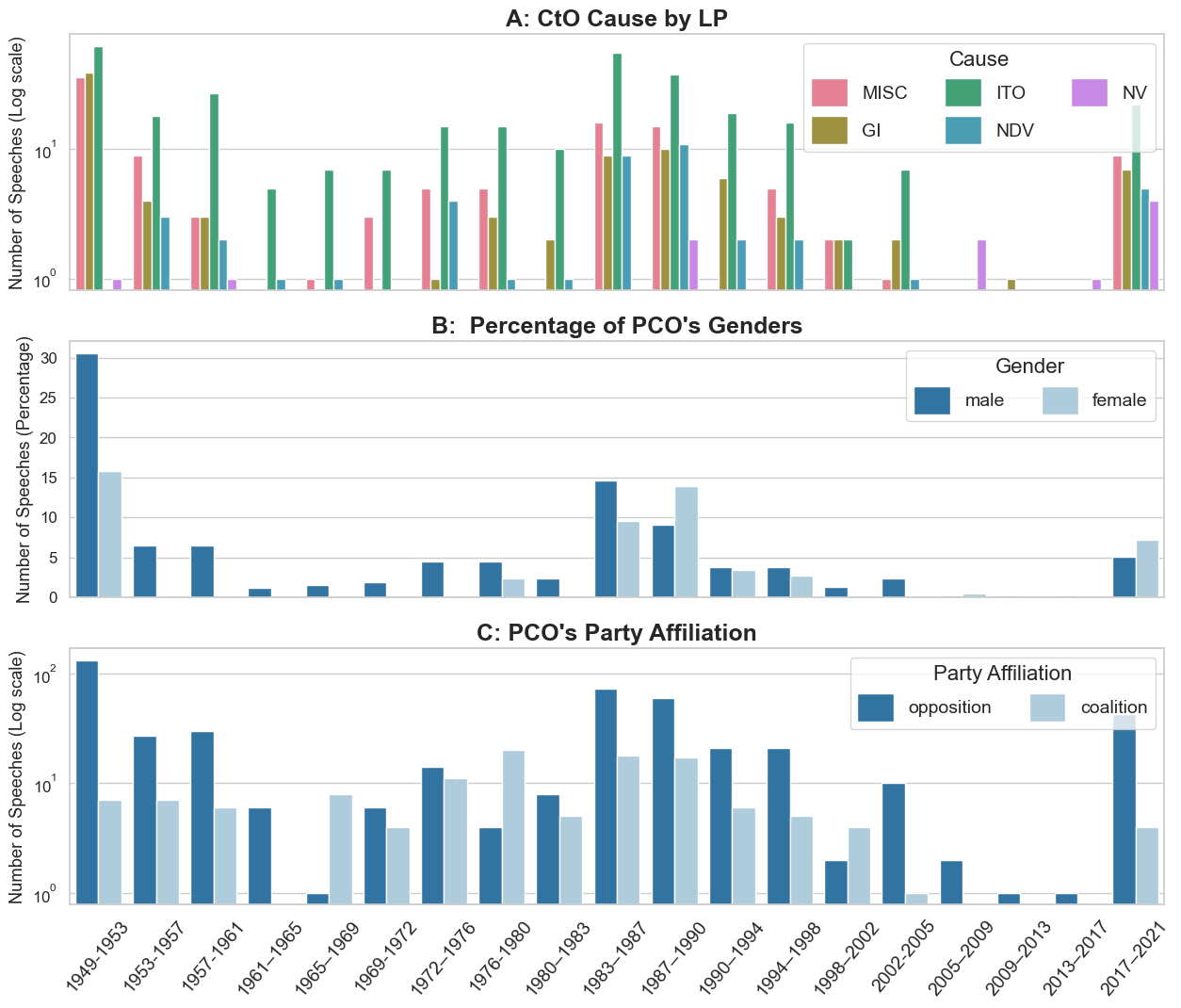}
  \caption{Distribution of causes, genders and party affiliations of PCOs over legislative periods (LPs). }
  \label{fig:freq_misc}
\end{figure}

The analysis revealed that ITO is the most prevalent trigger for a CtO, with a median occurrence of 17 per legislative period (LP), followed by MISC (median of 6 per LP), GI (median of 3 per LP), and NV (median of 1 per LP). Additionally, a total of 48 instances of NDV were identified but were either not transcribed or could not be located within our dataset (Table~\ref{tab:causes_number}).
As illustrated in Figure~\ref{fig:freq_misc}-A, the distribution of causes across LPs is non-uniform, with high standard deviations observed for all causes. The most frequently occurring cause, ITO, is present in all LPs except for LPs 16, 17, and 18. 

The $\chi^2$ test for independence was conducted using the Monte Carlo method to assess relationships between variables, as most of the data did not meet the assumptions required for the $\chi^2$ test. To assess the association strength between variables, we additionally applied Cram\'er’s V measure using the $\chi^2$ statistics from the Monte Carlo simulation\footnote{For the analysis of the relationship between party affiliation, gender, the name and PCO's party affiliation, only disambiguated individuals were considered.}. 
As Table~\ref{tab:stat_res2} demonstrates, statistically significant relationships were found between a CtO cause and LP, date,  year, and the session's sequence number in the LP. However, the associations between these variables were negligible. 

Figure~\ref{fig:freq_misc}-B demonstrates that, overall, men are called to order more frequently than women. However, in LPs 11, 16, and 19, a greater proportion of female PMs were called to order compared to their male counterparts. 
The median number of men receiving a call to order per LP is 19, compared to 5.5 for women. This corresponds to 3.72\% of all male parliamentarians, while the proportion for women is close to zero. However, high standard deviations of 31.99 and 6.57, respectively, were observed, indicating substantial variability in the data. Overall,  5.25\% of men and 2.62\% of women were called to order through the history of the German parliament (Appendix~\ref{app:data_stat}, Table~\ref{tab:gender_number_app}).
Statistically significant relationships were found between the gender of a person called to order (PCO) and the cause of the CtO, LP, and the session's president. Additionally, a moderate association was observed between the session's president and the PCO's gender, as well as between the PCO's gender and the LP. In contrast, a weak association was found between PCO's gender and the cause of the CtO. 

\begin{table}[h]
\scriptsize
    \centering
    \begin{tabular}{p{.05\textwidth}p{.07\textwidth}p{.08\textwidth}p{.06\textwidth}p{.07\textwidth}}
    \hline
       &   &  total frequency & median per LP & standard deviation\\
       \hline
       \parbox[t]{2mm}{\multirow{6}{*}{\rotatebox[origin=c]{90}{cause}}} & ITO  & 344 & 17 & 18.80\\
       & MISC & 120 & 6 & 10.54\\
       & GI  & 106 & 3 & 11.20\\
       & NDV  & 48 & 2 & 3.20\\
       & NV  & 13 & 1 & 1.46\\
       \hline
        \multirow{2}{2mm}{party affiliation} & opposition  & 460 & 10 & 33.14\\
       & coalition & 123 & 6 & 5.72\\
        \hline
    \end{tabular}
    \caption{Total number of speeches containing a specific cause triggered CtO and total number of CtOs distinguished by PCO's party affiliation in the corpus, and the median frequency by LP.}
    \label{tab:causes_number}
\end{table}

As presented in Figure~\ref{fig:freq_misc}-C, opposition party members receive more CtOs than coalition party members, with a median of 10 per legislative period compared to 6 for coalition members. However, high standard deviations indicate significant variability in the data (Table~\ref{tab:causes_number}). A moderate association was found between the session president and the PCO’s party affiliation. Additionally, a strong association was observed between the gender of the session president and the PCO’s party, as well as between the president of the session and the PCO. In contrast, the president’s party showed only a weak association with the PCO’s party. 

\begin{figure}[h]
  \centering
  \includegraphics[width=1\linewidth]{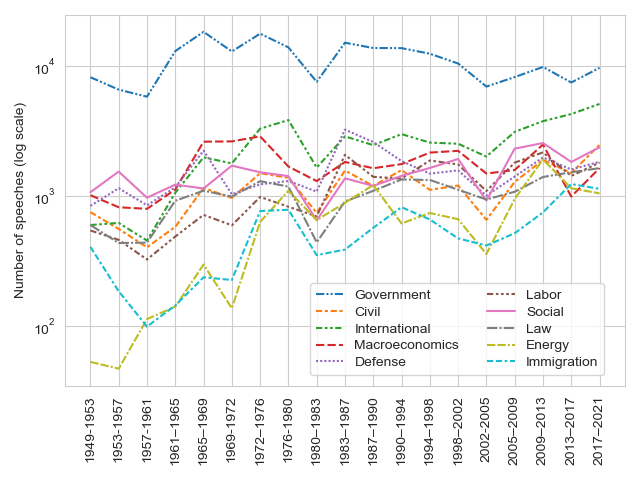}
  \caption{Distribution of the top 10 topics that caused CtOs over legislative periods. }
  \label{fig:top_topics}
\end{figure}

A statistically significant relationship was found between the discussed topic and the presence of a CtO in speech, though a negligible association was observed. The highest number of CtO causes was observed in speeches related to governmental issues (188), followed by presidency actions (89), civil affairs (56), and international affairs (48). No CtOs were recorded in discussions on foreign affairs and culture. Figure~\ref{fig:top_topics} illustrates the distribution of the top 10 topics containing CtOs over the 72 years. Government remains the most discussed topic across all LPs, and the number of speeches on most topics has increased over time. However, there was a sharp decline in immigration-related speeches between LPs 1 and 3, with a continued decrease in subsequent LPs. No statistically significant associations were found between the gender of the session president and the gender of the PCO's, the PCO's party affiliation, or the cause of the CtO. Additionally, no statistically significant relationship was found between the presence of CtO trigger in a speech and the speech's position (sequence number) in the agenda.

\begin{table}[h]
\tiny
    \centering
    \begin{tabular}{p{.1\textwidth}p{.133\textwidth}p{.05\textwidth}p{.03\textwidth}p{.04\textwidth}}
\hline
variable1  & variable 2 & $\chi^2$ statistics & p-value & Cram\'er's V \\
\hline
\multirow{5}{4em}{name of the president} & name of the PCO & 17671.519 & 0.000 & \textbf{0.795} \\
 & gender of the PCO  & 124.266 & 0.000 & \textbf{\textit{0.462}} \\
 & party of the PCO  & 1628.249 & 0.000 & \textit{\textbf{0.464}} \\
 & cause of the CtO  & 373.997 & 0.000 & \textbf{\textit{0.400}} \\
 & PCO's party affiliation  & 159.944 & 0.000 & \textbf{0.524} \\
\hline
\multirow{4}{4em}{gender of the president} & gender of the PCO  & 0.003 & \textit{1.000} &  \\
 & party of the PCO  & 62.008	 & 0.000 & \textbf{\textit{0.326}} \\
 & cause of the CtO  & 7.643 & \textit{0.100} &  \\
 & PCO's party affiliation  & 2.998	 & \textit{0.108} &  \\
\hline
party of the president & party of the PCO  & 273.806 & 0.000 & \textit{0.280} \\
\hline
name of the PCO & cause of the CtO  & 1186.054	 & 0.000 & \textbf{0.713} \\
\hline
\multirow{2}{5em}{gender of the PCO} & cause of the CtO  & 9.860 & 0.043 & \textit{0.130} \\
 & legislative period  & 93.080 & 0.000 & \textbf{\textit{0.400}} \\
\hline
party of the PCO & cause of the CtO  & 165.645 & 0.000 & \textit{0.267} \\
\hline
\hline
\multirow{6}{4em}{CtO trigger} & date  & 11475.096	 & 0.000 & \textit{0.109} \\
 & LP  & 1172.845 & 0.000 & 0.035 \\
 & session's sequence number in LP  & 762.647 & 0.000 & 0.028 \\
 & speech's sequence number in the agenda  & 994.335 & \textit{0.051} &  \\
 & discussed topic  & 370.055 & 0.000 & 0.020 \\
 & year  & 1385.469 & 0.000 & 0.038 \\
\hline
\end{tabular}
    \caption{The $\chi^2$ test with a Monte Carlo method and Cram\'er's V. P-value above the threshold marked with italicized text. A small association is marked with italicized text, a medium association with italicized bold text, and a large association with bold text. Sample size: 583 for variables above the double line; 958,100 for all other variables.}
    \label{tab:stat_res2}
\end{table}

\begin{comment}
\begin{figure}[h]
  \centering
  \includegraphics[width=\linewidth]{diagrams/topics_cto.png}
  \caption{Distribution of topics and CtO inclusion in the topic.}
  \label{fig:cto_topics}
\end{figure}
\end{comment}

\section{Conclusion}\label{sec:conclusion}

%{RQ2:} What are the underlying reasons for issuing calls to order?
In this study, we conducted a manual analysis of CtOs in the German parliament and developed a classification system comprising five underlying reasons for their issuance. Our analysis indicates that ITO (insult towards an individual) is the most frequent trigger of CtO. NV (non-verbal) accounts for the smallest share of CtO triggers. Moreover, statistical testing suggests that certain presidents are more likely to issue CtOs in response to specific types of triggers. Additionally, particular parliamentary members tend to employ specific classes of insults. At the same time, no gender-specific classes of insult were observed.

%{RQ1:} Which topics caused most calls to order?
Following, we applied a classification model that distinguished speech contributions to 21 topics, including presidency actions, as an additional category. Most CtO triggers were detected in speeches dedicated to governmental issues and presidency actions. The $\chi^2$ test revealed a statistically significant association between the topic and the CtO trigger; however, a Cram\'er's V showed only a weak association between these variables, which indicates that this association is not of practical interest and might occur due to the large data sample size. 
    
%{RQ3:} How do factors such as political party affiliation, individual politicians, legislative periods, and topics relate to issuing calls to order?
%Further statistical analysis uncovered several interesting trends. 
Notably, session presidents tend to call particular individuals to order preferentially. Moreover, presidents are more likely to call representatives of certain parties and genders to order. In addition, CtOs are associated with the party's affiliation. Generally, male individuals and opposition party members receive more calls to order than their female and coalition party counterparts. This supports the hypothesis that opposition members are more prone to breaching parliamentary order. Historically, there are fewer women than men in the German parliament, which can contribute to the pattern. Furthermore, the likelihood of being called to order varies by gender, depending on LP. However, no statistically significant relationship was detected between the gender of the session president and that of the PCO. 

In conclusion, both statistical and empirical evidence suggest that despite strict regulations, issuing CtOs is often subjective and significantly influenced by the session president and prevailing parliamentary trends.

%\clearpage
\section{Limitations}

This study is subject to several limitations. Firstly, we employed a semi-automated method to annotate the corpus. Sentences containing CtO instances were identified using a rule-based approach, which is a legitimate choice in this context, given that disciplinary measures in the German parliament are strictly regulated and, therefore, exhibit specific patterns. Nevertheless, a manual review revealed that these patterns occasionally resulted in false positives, as illustrated in the following example: 
\begin{itemize}
    \item \textbf{DE: }\textit{Ich kann nur wegen der Zwischenrufe zur Ordnung rufen, die ich selber höre.}
    \item \textbf{EN: }\textit{I can only call to order the interjections that I hear myself.}
\end{itemize}
Furthermore, the rule-based approach may not detect CtOs issued using non-conventional phrasing if such occurs in the dataset. We opted against the approach proposed by \citet{Jenny_Marcelo_2021}, as it showed a correct prediction rate of only 75.3\%, and we believe that this would not capture CtOs triggered by speeches lacking explicit negative connotations, as demonstrated in the following example:

\begin{itemize}
    \item \textbf{DE: }\textit{ Die Oder-Neiße-Grenze ist die Grenze des Friedens.}
    \item\textbf{EN: }\textit{The Oder-Neisse border is the border of peace.}
\end{itemize}

Secondly, a semi-automated approach was utilized to extract and disambiguate called-to-order individuals, which also may lead to false annotations.
%In addition, the analysis of the data and development of classification, as well as corresponding manual annotation of CtOs, was conducted by a single annotator, which introduces the potential for subjective bias. 

%However, the current research aims to provide an initial analysis of the data; to the best of our knowledge, there is no existing classification or in-depth analysis of calls to order. 
%To enhance the reliability of future annotations, we plan to engage two more annotators to produce a new version of the dataset for future research, e.g. for machine learning tasks. 

Additionally, for different reasons, we were not able to disambiguate all individuals mentioned in CtOs, nor all speakers in the corpus. Therefore, some statistical tests were conducted only with the disambiguated data. A total of 96 speeches containing CtO could not be disambiguated.

Moreover, we used a classification model \cite{klamm2022frameast} to find discussed topics in the speeches. This model was specifically trained to distinguish topics in speeches in the German parliament. However, the F1-score for some categories, such as Social Welfare and Public Lands, was under 0.5, which can cause false classification of speeches containing this topic. We decided to apply the model to the whole speech text and not to single sentences or paragraphs, as generally, one speech is dedicated to a specific topic. There could be some variations from the topic during the speech due to interjections, but the general topic stays the same. For future work, we also consider applying other techniques, such as the seeded Latent Dirichlet allocation as in \citet{doi:10.1177/0894439320907027}.

In addition, this paper focuses on the first analysis of calls to order; to our knowledge, there is no equivalent research in this area. Therefore, we focused on a general analysis of factors influencing incivility in parliamentary debates without a deeper investigation of single factors.  

Notably, in our statistical analysis, we accounted for the uneven distribution of some variables (such as gender, occurrence of a CtO in speech or topics) in our dataset and adjusted the statistical analysis accordingly. All the reported findings are statistically significant. 

Moreover, to the best of our knowledge, no pre-existing classification frameworks for calls to order currently exist. This paper marks the first attempt to systematically analyze and categorize such calls to order. While calls to order can be classified in various ways, such as by focusing on specific insult types, this study emphasizes overarching features of insulting behaviour. 

Finally, due to the absence of a benchmark dataset for this task, a quantitative evaluation of the rule-based methods was not feasible. However, because of the limited size of the analyzed dataset, all rule-based annotations were verified manually.

\section{Acknowledgments}
All authors were funded by the European Union under the Horizon Europe grant OMINO (grant number 101086321, \cite{holyst2024}). Views and opinions expressed are, however, those of the authors only and do not necessarily reflect those of the European Union or the European Research Executive Agency. Neither the European Union nor the European Research Executive Agency can be held responsible for them. Nina Smirnova additionally received funding from the Deutsche Forschungsgemeinschaft (DFG) under grant number: MA 3964/7-3 (POLLUX Project). 
%All sources of funding and contributions from collaborators will be acknowledged after acceptance.
\section{Bibliographical References}\label{sec:reference}

\bibliographystyle{lrec2026-natbib}
\bibliography{lrec2026-example}

\section{Language Resource References}
\label{lr:ref}
\bibliographystylelanguageresource{lrec2026-natbib}
\bibliographylanguageresource{languageresource}

\appendix
\onecolumn

\section{Classification schema}\label{app:annot_schema}

Based on the manual review of the dataset, we propose the following classification of actions that caused a call to order (Table~\ref{tab:classification_ext}). 

\begin{table}[h]
\small
    \centering
    \begin{tabular}{p{.12\textwidth}p{.1\textwidth}p{.16\textwidth}p{.26\textwidth}p{.26\textwidth}}
    \hline
        \textbf{class name} & \textbf{abbreviation} & \textbf{description} & \textbf{example DE} & \textbf{example EN}\\
    \hline    
    insult towards individual  & \textbf{ITO} & insult towards an individual & Schreiner [SPD]: Wild gewordener Gartenzwerg! & Schreiner [SPD]: Garden gnome gone wild!\\
    \hline
    general insult  & \textbf{GI} & insult towards a group of people, party, event, actions, etc. & Abg. Renner: Die Union der Faschisten von gestern ist fertig!&  PM Renner: The Union of Fascists of yesterday is finished!\\
    \hline
    non-verbal   & \textbf{NV}  & non-verbal actions that caused a call to order & Abgeordnete der Fraktion Die Linke halten Transparente und Fahnen hoch. & Members of the parliamentary group Die Linke hold up banners and flags.\\
    \hline
    not documented verbal & \textbf{NDV} & verbal actions that caused a call to order but were not transcribed. & Der Abg. Dr. Richter [Niedersachsen] wendet sich dem amtierenden Präsidenten zu und spricht unter andauernder großer Unruhe des Hauses auf ihn ein, ohne daß seine Worte vom Haus und am Stenographentisch verstanden werden können. & PM Dr Richter [Lower Saxony] turns to the President-in-Office and speaks to him, to the continued great agitation of the House, without his words being understood by the House and the stenographers' table.\\
    \hline
    miscellaneous  & \textbf{MISC} &  all other verbal actions excluding direct insults that caused a call to order & Gerd Andres [SPD]: Wie lange darf der eigentlich noch reden, Herr Präsident? Ist das unbegrenzt? &  Gerd Andres [SPD]: How long is he actually allowed to talk, Mr President? Is that unlimited?\\
    \hline
    \end{tabular}
    \caption{Classification schema}
    \label{tab:classification_ext}
\end{table}

\section{Dataset statistics}\label{app:data_stat}
\begin{table}[H]
\small
\centering
    \begin{tabular}{ll}
    \hline
       & count \\
    \hline
    number of speeches containing a CtO with disambiguated individuals called to order  &  493 \\
    number of speeches containing a CtO with not disambiguated individuals called to order  &  65 \\
    \hline
    \end{tabular}
    \caption{Number of speech contributions and calls to order in GermaParl corpus}
    \label{tab:corpus_stat_app}
\end{table}

\begin{table}[H]
\small
    \centering
    \begin{tabular}{p{.07\textwidth}p{.11\textwidth}p{.11\textwidth}p{.1\textwidth}p{.12\textwidth}p{.1\textwidth}p{.12\textwidth}p{.1\textwidth}}
    \hline
       PCO gender&  number of PMs called to order  &  \% of PMs called to order&number of PMs  in parliamnet& median number of PMs called to order per LP& standard deviation &median \% of PMs called to order per LP&standard deviation (\%)\\
       \hline
        male  & 493 & 5.25 & 9390& 19 & 31.99 &3.72 &	7.07\\
        female & 59  & 2.62 & 2249 & 5.5 & 6.57 &0 &4.99 \\
        \hline
    \end{tabular}
    \caption{Number of PCOs distinguished by their gender.}
    \label{tab:gender_number_app}
\end{table}

\end{document}